\documentclass[runningheads]{llncs}
\usepackage[T1]{fontenc}
\usepackage{graphicx}
\usepackage{cite}
\usepackage{multirow}
\usepackage{amsmath,amssymb,amsfonts}
\usepackage{algorithmic}
\usepackage{graphicx}
\usepackage{textcomp}
\usepackage{xcolor}
\usepackage{graphicx}
\usepackage{subfigure}
\usepackage{wrapfig}
\usepackage{algorithm, algorithmic}
\usepackage{color, colortbl}
\usepackage{multirow}
\usepackage{listings}
\usepackage{soul}
\definecolor{Gray}{gray}{0.9}

\input{Definitions.tex}

\graphicspath{{figures/}}

\def\BibTeX{{\rm B\kern-.05em{\sc i\kern-.025em b}\kern-.08em
    T\kern-.1667em\lower.7ex\hbox{E}\kern-.125emX}}
\usepackage{wrapfig}

\begin{document}
\title{FedReview: Review and Dispose Poisoned Updates without Validation Datasets or Historic Knowledge}
\titlerunning{FedReview: Review and Dispose Poisoned Updates}
\author{Tianhang Zheng\inst{1,2}\thanks{Corresponding author: Tianhang
Zheng <\email{zthzheng@zju.edu.cn}>.} \and
Yanlu Li\inst{1,2} \and
Bohan Deng\inst{3} \and
Baochun Li\inst{4}}
\authorrunning{T. Zheng et al.}
\institute{
The State Key Laboratory of Blockchain and Data Security, Zhejiang University, Hangzhou, China\\
\and
Hangzhou High-Tech Zone (Binjiang) Institute of Blockchain and Data Security, Hangzhou, China\\
\email{\{zthzheng,liyanlu\}@zju.edu.cn}
\and
ZJU-UIUC Institute, Zhejiang University, Hangzhou, China\\
\email{bohan1.23@intl.zju.edu.cn}
\and
University of Toronto, Toronto, Canada\\
\email{bli@ece.toronto.edu}
}
%\author{Anonymous submission}

%
\maketitle              % typeset the header of the contribution
\begin{abstract}
Federated learning has emerged as a decentralized approach for training high-performance models without accessing user data. Despite its effectiveness, it is vulnerable to poisoning attacks, where malicious users manipulate the global model by uploading poisoned updates. In this paper, we propose FedReview, a review-based mechanism to identify and dispose the potential poisoned updates in federated learning. Under FedReview, the server randomly assigns a subset of clients as reviewers to evaluate model updates on their training datasets in each round. The reviewers rank the updates based on evaluation results and estimate the number of low-quality updates as potential poisoned ones. Based on the review reports, the server applies a majority voting mechanism to aggregate rankings, which tolerates wrong rankings from malicious reviewers and guides the removal of suspicious updates during model aggregation. In contrast to prior works such as FLTrust, FedReview does not require a server-side validation dataset or prior knowledge of clients, allowing flexible client participation. Extensive experiments demonstrate that FedReview enables the server to learn a well-performing global model in adversarial environments.
\end{abstract}
\section{Introduction}
\label{sec:intro}
\vspace{-7pt}
Over the past few years, deep learning has made a series of substantial breakthroughs due to the availability of massive training data. In spite of those impressive breakthroughs, the widespread application of deep learning in the real world is still facing a variety of challenges. One imperative challenge is the concern from many users about sharing their sensitive data for training deep learning models. To overcome this challenge, the community proposed a decentralized learning technique called federated learning. Federated learning enables the users to train models on their local devices and involves a server to aggregate the training results for updating a global model.
Therefore, federated learning does not require direct access to the user data to train deep learning models.

While federated learning attempts to safeguard user data, it simultaneously introduces a critical attack vector, known as model poisoning. Model poisoning occurs when an adversary, either hiding among the users or compromising some user devices, corrupts the model by uploading poisoned model updates to the server. This new attack vector has sparked significant research effort within the community to investigate new model poisoning attacks and defenses.

On one hand, the community has proposed several model poisoning methods \cite{bagdasaryan2020backdoor,fang2020local, cao2021fltrust} to facilitate the exploration of the risks raised by model poisoning in different scenarios. On the other hand, several defensive methods have been developed against model poisoning, such as robust aggregation methods \cite{blanchard2017machine, yin2018byzantine}, which compute a robust estimation of the averaged update over the benign and poisoned updates, mitigating their negative effects.
% to mitigate the negative effects of poisoned updates. 
To circumvent those robust aggregation methods, some recent works \cite{fang2020local, shejwalkarmanipulating} further developed adaptive model poisoning attacks to generate poisoned updates that can bypass the criterion of those robust aggregation methods. Notably, the attacks proposed by \cite{shejwalkarmanipulating}, such as min-max and min-sum attacks, significantly reduce the accuracy of federated learning, even under protection of robust aggregation.

Despite the remarkable effectiveness of min-max and min-sum attacks against robust aggregation methods, 
% we observe that 
most previous works \cite{shejwalkarmanipulating, zhang2023oblivion} evaluate these attacks under a special setting, where the users upload model gradients or single-epoch model updates. 
In practical scenarios, if the participating clients learn model updates via multi-epoch local training ({\em e.g., 5 epochs}), we find that min-max and min-sum attacks can not cause severe performance degradation. Through extensive analysis, 
we demonstrate that min-sum and min-max attacks using the inverse unit vector as the perturbation vector are equivalent to the scaling model poisoning attack \cite{bagdasaryan2020backdoor} with a dynamic scaling factor, which is too small to induce severe negative impacts on the global model. Increasing this scaling factor to an appropriate value leads to a substantial reduction in the model accuracy, even if the server applies robust aggregation methods.

Since robust aggregation methods are still vulnerable to model poisoning, the community has proposed several advanced defenses, such as FLTrust \cite{cao2021fltrust} and FLDetector \cite{zhang2022fldetector}. In contrast to robust aggregation methods, FLTrust and FLDetector can detect the majority of malicious clients. Nevertheless, FLTrust requires the server to possess a clean validation dataset with a distribution similar to the user data distribution. According to \cite{zhang2022fldetector}, FLTrust exhibits poor performance when the distribution of the validation dataset deviates from the user data distribution. FLDetector has a prerequisite about high consistency between the current and historic model updates from a benign client. Therefore, FLDetector requires the server to have historic knowledge about clients and cannot handle the situation where the clients are allowed to drop out and rejoin with different identification numbers.

Different from the previous defensive methodologies, we introduce a distributed review mechanism called FedReview, \emph{which is free from prerequisites of FLTrust and FLDetector ({\em i.e.,} validation datasets and consistency between current and historic updates)}, to identify and dispose potential poisoned updates in federated learning. Under FedReview, the server needs to randomly select a subset of clients as reviewers to evaluate the model updates and submit review reports. Each review report comprises two components---an estimated number of poisoned updates and a ranking of the model updates. After collecting the reviews, the server can obtain a reliable estimation of the number of potential poisoned updates. Based on the collected rankings, the server can further leverage a simple but effective majority vote mechanism to obtain the indices of the potential poisoned updates. 

We conduct a comprehensive set of experiments to evaluate FedReview on Purchase-100, EMNIST, CIFAR-10, and FEMNIST. We demonstrate that FedReview can correctly identify the poisoned updates with high precision. We compare FedReview with multiple robust aggregation methods, including M-Krum, Trimmed Mean, and Median, which also do not require the server to possess any validation data or historic knowledge. Our evaluation results indicate that FedReview outperforms those methods by up to $30\%$ in model accuracy.
\vspace{-10pt}
\section{Preliminaries}\label{sec:preliminary}
\vspace{-8pt}
\paragraph{Definitions and Notations}
We denote a data sample by $\bm{x}$ and its label by $y$. We denote the label set by $\mathcal{Y}_y = \{1, 2, ..., \mathcal{Y}\}$ with totally $\mathcal{Y}$ labels. We represent a neural network by $\bm{f}_{\bm{\theta}}(\cdot)$ with model weights $\bm{\theta}$. $\bm{f}_{\bm{\theta}}(\bm{x})$ refers to the softmax output of $\bm{x}$, and $\ell(\bm{f}_{\bm{\theta}}(\bm{x}), y)$ refers to the cross-entropy between $\bm{f}_{\bm{\theta}}(\bm{x})$ and $y$. i.i.d. is the abbreviation of independent and identically distributed. In terms of the hyperparameters of federated learning, we denote the total number of training rounds by $T$ and the set of clients by $\mathcal{S}$. We represent the number of selected clients in each round by indices $\{1, 2..., C\}$. We denote the $c$-th client's training dataset by $D_c$. 

\vspace{-10pt}
\paragraph{Federated Learning}
Federated Learning (FL) is proposed as a decentralized learning technique for data privacy protection~\cite{yang2019federated, li2020federated}. A general setup of federated learning needs a server to coordinate a number of clients for the purpose of optimizing a global model through multiple-round training and communication.
In each training round, the server first selects several clients and sends the current global model weights to the selected clients. Sequentially, the selected clients train the received model on their local training datasets, and upload the updated local models back to the server. Finally, the server aggregates the local models to update the global model and starts a new round.
The most common aggregation method is FedAvg \cite{mcmahan2017communication}.

\vspace{-10pt}
\paragraph{Model Poisoning}
The setup of federated learning provides a malicious client a chance to manipulate the global model by poisoning the uploaded model updates. This direct manipulation on model weights by poisoned model updates significantly enhances the effectiveness of poisoning attacks, compared to the indirect impact of data poisoning \cite{tolpegin2020data} on the model weights. To generate the poisoned updates, the adversary can learn the updates using a contaminated dataset and scale up the updates to amplify their effects \cite{bagdasaryan2020backdoor}. However, advanced Byzantine-robust aggregation algorithms mentioned in  Section~\ref{subsec:byzantine} can substantially mitigate the effects of the poisoned updates.
  
\vspace{-10pt}
\paragraph{Byzantine-Robust Aggregation}\label{subsec:byzantine}
Byzantine-robust aggregation methods are widely used to defend against model poisoning attacks, particularly when the server does not possess a validation dataset to evaluate client updates. Representative methods include Multi-Krum \cite{blanchard2017machine}, Trimmed Mean\cite{yin2018byzantine}, and Median \cite{yin2018byzantine}, which aim to reduce the influence of malicious updates by aggregating client models. In this paper, we implement the above three methods for comparison with our defensive mechanism.

\vspace{-15pt}
\section{Threat Model}\label{sec:threat}
\vspace{-5pt}
\subsection{Adversary Assumptions}

\paragraph{Adversary's Objective}
In this paper, the adversary's objective is to decrease the accuracy of the global model, which is similar to the adversary's goal in \cite{fang2020local, shejwalkarmanipulating}. The attacks driven by this adversary's objective are called untargeted model poisoning attacks. In contrast to \cite{shejwalkarmanipulating}, the adversary studied in the paper crafts poisoned model updates rather than malicious gradients, and upload the poisoned updates with the server to achieve the attack goal.

\vspace{-8pt}
\paragraph{Adversary's Knowledge}\label{subsec:adv_knowledge}
Since this paper presents a review mechanism to defend against model poisoning, we mainly consider a strong adversary with the knowledge of the benign devices for evaluation. When the model updates are sent from the users to the server through unencrypted channels, or all the updates are encrypted with a shared secret key, the adversary is able to know the updates from the benign devices. If the model updates are encrypted with user-specific secret keys, the adversary may not know the updates. But in that case, key management could be challenging and costly since federated learning usually involves a large number of users in the training stage.

In terms of knowledge about the defense method, we consider two cases: (1) The adversary knows and leverages the defense method to design an adaptive attack; (2) The adversary does not use the defense method in its attack. For the first case, the adversary could adopt the adaptive attack. For the second case, the adversary could adopt the model poisoning attacks introduced in Section~\ref{subsec:naive_attack}~\&~\ref{subsec:optim_model_poison}.

\vspace{-8pt}
\paragraph{Adversary's Capabilities}\label{subsec:adv_capability}
Following the previous literature, we consider that the adversary is able to control $20\%$ of the clients by default. If the server select any client from the controlled $20\%$ clients, the client will upload a poisoned update provided by the adversary using the attacks in Section~\ref{sec:attack}. In the experiments, we also consider other settings of the proportion of the compromised clients. The adversary is also capable of choosing an appropriate attack method based on its knowledge, as introduced in Section~\ref{subsec:adv_knowledge}.
%  If the server select any client from the controlled $20\%$ clients, the client will upload a poisoned update, which is provided by the adversary using the attacks in Section~\ref{sec:attack}, to the server.
% appropriate attack method according to its knowledge, as introduced in Section~\ref{subsec:adv_knowledge}.

\vspace{-10pt}
\subsection{Other Assumptions}
\vspace{-5pt}

We assume that the server has the same capabilities as in the standard setup of federated learning, which means \ul{the server does not have a validation dataset} that can represent the properties of the user data distribution. Besides, we assume that the server does not have historic knowledge, meaning that \ul{the server cannot rely on the consistency between historic and current updates from one client for defending against model poisoning. This assumption allows all the clients to drop out and rejoin the federated learning process with different identification numbers.} \textit{Given these assumptions, we do not include FLDetector and FLTrust in the baselines for comparison.}

\vspace{-8pt}
\section{Model Poisoning Attacks}\label{sec:attack}
\vspace{-5pt}
\subsection{Scaling Model Poisoning Attack}\label{subsec:naive_attack}
\vspace{-5pt}
Untargeted model poisoning attempts to degrade the model performance by uploading poisoned updates to the server. Given this adversary goal, we could formulate a scaling model poisoning attack to craft the poisoned update, {\em i.e.,}
\vspace{-5pt}
\begin{equation}
    \scalebox{0.85}{$
        \displaystyle
        \Delta\thetab^t_m  = -\lambda \Delta\thetab^t~~~
	s.t.~~\Delta\thetab^t  = \frac{1}{|\mathcal{S}_t|}\sum_{i \in \mathcal{S}_t}\Delta\thetab^t_i,
    $}
\end{equation}
%\vspace{-3pt}
where $\mathcal{S}_t$ refers to the subset of clients selected for the $t$-th round. $\Delta\thetab^t_i$ denotes the multi-epoch updates from the selected clients. 
$\Delta\thetab^t$ is the average of the client updates, adopted as global model update. $\Delta\thetab^t_m $ is the poisoned update introduced by the scaling attack, which is the opposite of the global model update scaled by a factor $\lambda$.
\vspace{-15pt}
\begin{figure}
\centering
\includegraphics[width=0.35\linewidth]{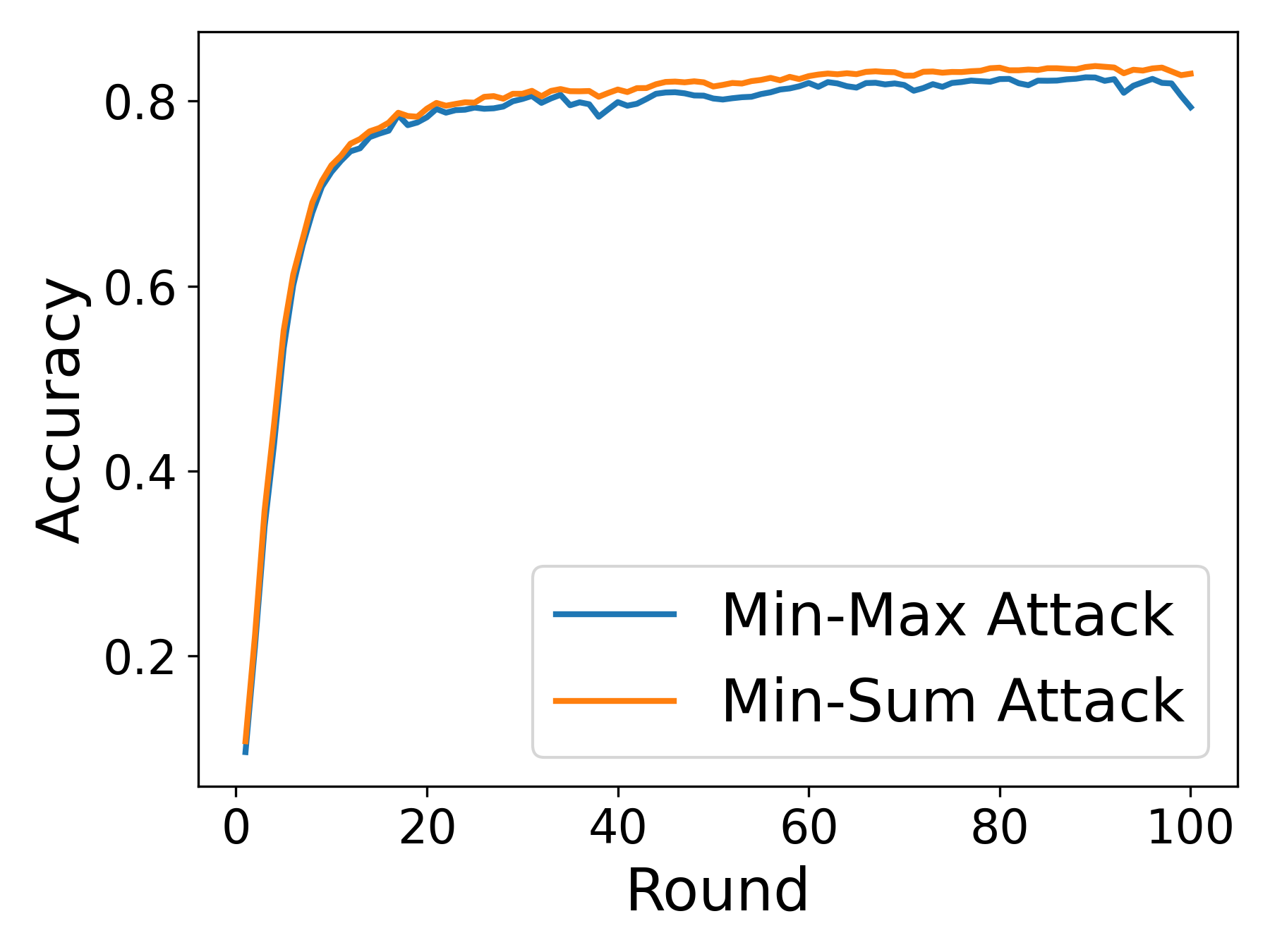} % Replace with your image file
  % \hfill % Space between the figures. Remove if you want the figures closer to each other
    \includegraphics[width=0.35\linewidth]{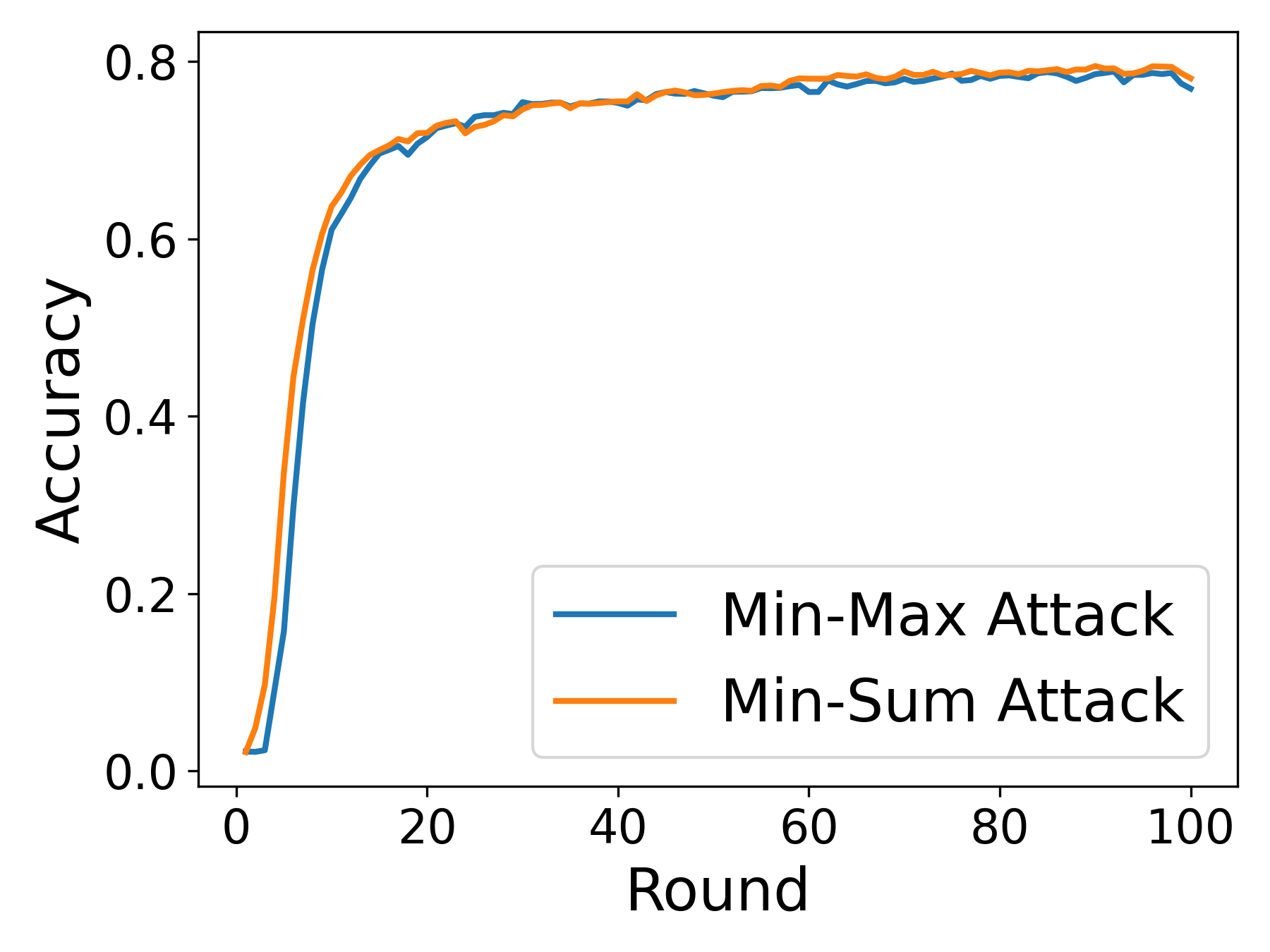} % Replace with your image file
  \vspace{-0.5cm}
  \caption{The test accuracy of FedAvg against the min-max attack. Left: Purchase-100, Right: EMNIST}
  \label{fig:min_max}
  \vspace{-0.6cm}
\end{figure}

Given the above formulation,
$\Delta\thetab^t_m $  can push the global model towards the opposite direction of the averaged benign update, {\em i.e.,} $\Delta\thetab^t$. $\lambda$ is a scaling factor to amplify the poisoned update. By default, the adversary could set $\lambda$ as the ratio of benign clients to malicious clients so that the positive effect of the benign updates will be neutralized by the negative effect of the poisoned updates.

\vspace{-12pt}
\subsection{Optimization based Model Poisoning}\label{subsec:optim_model_poison}

To bypass robust aggregation methods, the community has proposed several optimization based model poisoning methods, such as min-max and min-sum attacks in \cite{shejwalkarmanipulating}.
Since min-max and min-sum attacks are common benchmarks in existing literature \cite{zhang2023oblivion}, we detail their formulations below.
The min-max attack can be mathematically expressed as

\vspace{-10pt}
\begin{equation}\label{eq:min_max}
    \resizebox{0.9\linewidth}{!}{$
        \displaystyle
        \argmax_{\gamma} \max_{i \in \mathcal{S}_t}\|\Delta\thetab^t_m - \Delta\thetab^t_i\|_2 \leq \max_{i, j\in \mathcal{S}_t}\|\Delta\thetab^t_i - \Delta\thetab^t_j\|_2 ~~~ s.t.~~\Delta\thetab^t_m = \frac{1}{|\mathcal{S}_t|}\sum_{i \in \mathcal{S}_t}\Delta\thetab^t_i - \gamma \Delta\thetab^t_p,
    $}
\end{equation}
where $\frac{1}{|\mathcal{S}_t|}\sum_{i \in \mathcal{S}_t}\Delta\thetab^t_i$ is the mean of the benign updates $\Delta\thetab^t_i$, and $\Delta\thetab^t_p$ refers to a malicious update direction, which is usually set as the direction of $\frac{1}{|\mathcal{S}_t|}\sum_{i \in \mathcal{S}_t}\Delta\thetab^t_i$. The objective is to find the maximum $\gamma$ satisfying that the distance between the malicious update $\Delta\thetab^t_m$ and any benign update is smaller than the maximum distance between the benign updates. The malicious update obtained by optimizing Eq.~\ref{eq:min_max} is still close to the benign updates in terms of the Euclidean distance and thus may bypass robust aggregation methods.

The min-sum attack can be mathematically expressed as
\begin{equation}\label{eq:min_sum}
    \resizebox{0.9\linewidth}{!}{$
        \displaystyle
        \argmax_{\gamma} \sum_{i \in \mathcal{S}_t}\|\Delta\thetab^t_m - \Delta\thetab^t_i\|_2 \leq \sum_{i, j\in \mathcal{S}_t}\|\Delta\thetab^t_i - \Delta\thetab^t_j\|_2 ~~~ s.t.~~\Delta\thetab^t_m = \frac{1}{|\mathcal{S}_t|}\sum_{i \in \mathcal{S}_t}\Delta\thetab^t_i - \gamma \Delta\thetab^t_p,
    $}
\end{equation}
\vspace{-10pt}

whose objective is to find the maximum $\gamma$ satisfying that the sum of the distances between the malicious update   benign updates$\Delta\thetab^t_m$ and all benign updates is smaller than the sum of the distances between all benign updates. \emph{Although min-max and min-sum attacks seem to have complicated formulations, they are actually similar to the scaling attack but adopt an optimized dynamic scaling factor $\lambda$.} Specifically, the most commonly-used  $\Delta\thetab^t_p$ in Eq.~\ref{eq:min_max}~\&~\ref{eq:min_sum} is $\Delta\thetab^t/\|\Delta\thetab^t\|_2$, where $\Delta\thetab^t  = \frac{1}{|\mathcal{S}_t|}\sum_{i \in \mathcal{S}_t}\Delta\thetab^t_i$. As a result, the poisoned update in Eq.~\ref{eq:min_max}~\&~\ref{eq:min_sum}  can be rewritten as
\begin{equation}
    \scalebox{0.85}{$
        \displaystyle
        \Delta\thetab^t_m  = - (\frac{\gamma}{\|\Delta\thetab^t\|_2} - 1) \Delta\thetab^t
    $}
\end{equation}

\begin{wrapfigure}{L}{0.4\textwidth}
	\centering
    \vspace{-0.8cm}
	\includegraphics[width=\linewidth]{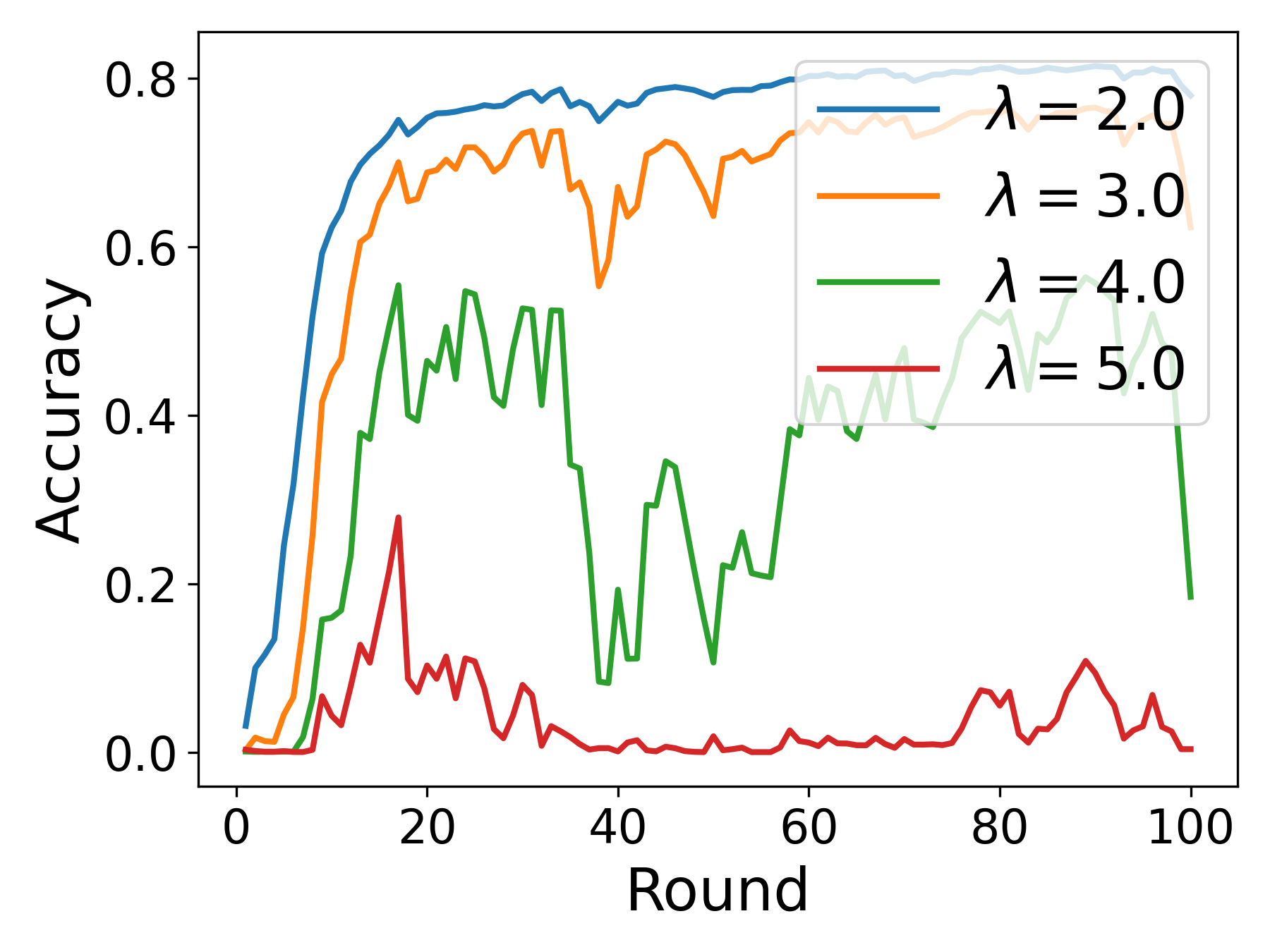}
    \vspace{-0.8cm}
	\caption{The testing accuracy of the global model under the scaling attack with different scaling factors.}
	\label{fig:scaling_factor}
    \vspace{-0.8cm}
\end{wrapfigure}

\emph{Therefore, min-max and min-sum attacks can be viewed as the scaling attack with a dynamic scaling factor $\lambda = \frac{\gamma}{\|\Delta\thetab^t\|_2} - 1$ in most cases.}
In the previous literature, min-max and min-sum attacks  are usually evaluated on the model gradients or the model updates from limited optimization steps. \emph{However, in practical federated learning, we find that min-max and min-sum attacks are ineffective, even if the server applies no defense.} Specifically, we conduct experiments on Purchase-100 and EMNIST, where the model updates are the results of 5-epoch local optimization. We report the testing accuracy in Fig.~\ref{fig:min_max} and observed that the min-max attack is ineffective.

By analyzing the learning process on Purchase-100, we observe that the dynamic scaling factor oscillates between $-1$ and $2$. As shown in Fig.~\ref{fig:scaling_factor}, experiments evaluating the scaling attack on Purchase-100 demonstrate that the scaling factor should exceed $2$ for effectiveness. Thus, the min-max attack is ineffective due to a small scaling factor. In another word, due to the small factor, the poisoned update of the min-max attack can not neutralize the positive effect of the benign updates in practical federated learning. As a result, in Section~\ref{sec:experiments}, we mainly employ the scaling attack with an appropriate factor to evaluate federated learning and the defenses.

\vspace{-12pt}
\section{FedReview Mechanism}\label{sec:defense}
\vspace{-10pt}
To defend against model poisoning under the threat model in Section~\ref{sec:threat}, we propose a review mechanism called FedReview to assess model updates and dispose the potential poisoned updates. The basic pipeline of FedReview for each training round is: The server first selects a subset of clients $\mathcal{S}_t$, and the clients in $\mathcal{S}_t$ are expected to upload their model updates to the server. Once receiving the updates, the server randomly selects another subset of clients from $\mathcal{S}/\mathcal{S}_t$ as reviewers and sends the updates to those reviewers for evaluation.

\vspace{-18pt}

\begin{algorithm}
	\caption{Review Mechanism for Federated Learning}
	\label{alg:review_mechanism}
    \scalebox{0.85}{
    \begin{minipage}{\linewidth}
	\begin{algorithmic}[1]
		\REQUIRE The set of clients $S$; total number of rounds $T$; global model $f_{\thetab}(\cdot)$ with model weights $\thetab$.
		\STATE  Initialize the model weights for $f_{\thetab}(\cdot)$, denoted by $\thetab^0$.
		\FOR{$t$ = $0$ to $T-1$}
		\STATE \textbf{Server}: Randomly select a subset of clients $S_t$ from $S$ and broadcast $\thetab^t$ to the selected clients.
		\STATE \textbf{Client $c$ ($c \in S_t$)}: Update or poison the global model weights, and send the update $\Delta \thetab_c^t$ to the server.
		\STATE \textbf{Server}: Receive updates from the clients and randomly select a subset of clients $R_t$ from $S/S_t$ as reviewers.
		\STATE \textbf{Server}: Send the updates $\{\Delta \thetab_c^t | c \in S_t\}$ to reviewers.
		\color{red}
		\STATE \textbf{Client (Reviewer) $r$ ($r \in R_t$)}: Evaluate $\{\thetab^t + \Delta\thetab_c^t | c \in S_t\}$ on its training data to obtain loss $\{loss_{c, r}^t | c \in S_t\}$.
		\STATE \textbf{Client (Reviewer) $r$ ($r \in R_t$)}: Estimate the number of adversaries, {\em i.e.,} $n_{adv}^r$, using Algorithm~\ref{alg:adv_estimation}.
		\STATE \textbf{Client (Reviewer) $r$ ($r \in R_t$)}: Rank the updates based on $\{loss_{c, r}^t | c \in S_t\}$.
		\color{black}
		\STATE \textbf{Client (Reviewer) $r$ ($r \in R_t$)}: Send the review reports including $n_{adv}^r$ and the ranking to the server.
		\color{blue}
		\STATE \textbf{Server}: Estimate the number of adversaries $n_{adv}$ by the median of $\{n_{adv}^r | r \in R_t\}$.
		\STATE \textbf{Server}: Aggregate the rankings from the reviewers and remove $n_{adv}$ updates based on majority vote.
		\color{black}
		\STATE \textbf{Server}: Aggregate the remaining updates to get $\thetab^{t+1}$.
		\ENDFOR
		\STATE Return $\thetab^{T}$.
	\end{algorithmic}
    \end{minipage}
}
\end{algorithm}
\vspace{-20pt}

FedReview selects reviewers from $\mathcal{S}/\mathcal{S}_t$ instead of $\mathcal{S}$, otherwise, a client may review its own update and produces a biased review.
The reviewers are requested to estimate the number of potential adversaries and rank the model updates. Sequentially, the reviewers include the estimated number and the rankings in their review reports and send the reports to the server.
Finally, the server aggregates the reviews to identify and remove the potential poisoned updates. 
We formulate the above pipeline as Algorithm~\ref{alg:review_mechanism}, where the \Red{red} part indicates how a reviewer creates a review report, and the \Blue{blue} part indicates how the server leverages the reviews to remove potential poisoned updates. In the following two subsections, we will detail how to create a review report and aggregate the reviews.

\vspace{-10pt}
\subsection{Review Components}
\vspace{-5pt}

As mentioned before, a review report contains two key components, {\em i.e.,} estimated number of poisoned updates (adversaries) $n_{adv}$ and rankings of model updates.
%We illustrate how to estimate $n_{adv}$ in Algorithm~\ref{alg:adv_estimation}.
To estimate $n_{adv}$, a reviewer $r$ first needs to evaluate the global model parameters plus the model updates from $\mathcal{S}_t$, {\em i.e.,} $\{\thetab^t + \Delta\thetab_c^t | c \in \mathcal{S}_t\}$, on its training dataset $D_r$ to compute the loss $\{loss_{c, r}^t | c \in S_t\}$. Formally, the reviewer $r$ computes the loss by
\vspace{-5pt}
\begin{equation}
    \scalebox{0.8}{$
        \displaystyle
        loss_{c, r}^t = \frac{1}{|D_r|}\sum_{(\bm{x}, y)\in D_r}\ell(\fb_{\thetab^t + \Delta\thetab_c^t}(\bm{x}), y).
    $}
\end{equation}
%\vspace{-0.2cm}
Our main intuition for estimating $n_{adv}$ is that the loss of poisoned model updates should be larger than the loss of benign updates, because the adversary aims to increase the loss and degrade the model performance. Thus, $n_{adv}$ should be the number of large outliers in $\{loss_{c, r}^t | c \in S_t\}$.
Note that under non-i.i.d. settings, the reviewers should use a class-balanced dataset sampled from $D_r$ instead of $D_r$ to compute $\{loss_{c, r}^t\}$ to avoid the biased evaluation.

\vspace{-15pt}
\begin{algorithm}
	\caption{Estimating Number of Poisoned Updates}
	\label{alg:adv_estimation}
    \scalebox{0.9}{
    \begin{minipage}{\linewidth}
	\begin{algorithmic}[1]
		\REQUIRE Loss of model weights $\{\thetab^t + \Delta\thetab_c^t | c \in S_t\}$ on the reviewer's training dataset, {\em i.e.,} $\{loss_{c, r}^t | c \in S_t\}$, a threshold $k$ (set as $1$ by default).
		\STATE Employ the median of $\{loss_{c, r}^t | c \in S_t\}$ as a robust mean $\mu_{loss}$.
		\STATE Employ $\sqrt{\textrm{median}(\{(loss_{c, r}^t - \mu_{loss})^2 | c \in S_t\})}$ as a robust standard deviation $\sigma_{loss}$.
  
		\STATE Count the number of $loss_{c, r}^t$ that satisfies $loss_{c, r}^t > \mu_{loss} + k \sigma_{loss}$, denoted by $n_{adv}^r$.
  
		\STATE Return the number $n_{adv}^r$.
	\end{algorithmic}
    \end{minipage}
}
\end{algorithm}
\vspace{-18pt}

We develop Algorithm~\ref{alg:adv_estimation} based on the above intuition. In Algorithm~\ref{alg:adv_estimation}, we employ the median of $\{loss_{c, r}^t | c \in S_t\}$ as a relatively robust estimation for the mean of the loss, which can avoid the negative impacts from large $loss_{c, r}^t$. Note that large $loss_{c, r}^t$ (loss on the poisoned updates) may significantly increase the arithmetic mean and leads to an underestimate of the number of outliers. 
Similarly, we employ $\sqrt{\textrm{median}(\{(loss_{c, r}^t - \mu_{loss})^2 | c \in S_t\})}$ as an estimation for the standard deviation of the loss. If $loss_{c, r}^t > \mu_{loss} + k\sigma_{loss}$, we consider $loss_{c, r}^t$ as an outlier and the corresponding update $\Delta\thetab_c^t$ as a potential poisoned update. Thus, we count the number of $loss_{c, r}^t$ in $\{loss_{c, r}^t | c \in S_t\}$ that satisfies $loss_{c, r}^t > \mu_{loss} + k\sigma_{loss}$ as the number of potential poisoned updates (potential adversaries).

To rank the model updates, the reviewer $r$ simply leverages the rankings of $\{loss_{c, r}^t | c \in S_t\}$. A large $loss_{c, r}$ indicate a top rank of the model update (large probability of being a poisoned update). After ranking the model updates, the reviewer $r$ can send its reviewer report to the server.

\vspace{-12pt}
\subsection{Review Aggregation}
%\vspace{-5pt}
%\vspace{-8pt}
\begin{algorithm}
	\caption{Majority Vote}
	\label{alg:majority_vote}
    \scalebox{0.9}{
    \begin{minipage}{\linewidth}
	\begin{algorithmic}[1]
		\REQUIRE Rankings of updates from all the reviewers $\{\bm{\pi}^r | r \in R_t\}$, where $\bm{\pi}^r(c)$ refers to the ranking of client $c$'s model update among all the updates. The reverse of $\bm{\pi}^r$ is $\bm{\omega}^r$, {
			\em i.e.,} $\bm{\omega}^r(\bm{\pi}^r(c)) = c$.
		\STATE Initialize a zero voting vector $\bm{v}$ with length $|\mathcal{S}_t|$.
		\FOR{$r \in R_t$}
		\FOR{$i = 0$ to $n_{adv}-1$}
		\STATE $\bm{v}[\bm{\omega}^r(i)] = \bm{v}[\bm{\omega}^r(i)]  + 1$
		\ENDFOR
		\ENDFOR
		\STATE Return $\{i | \bm{v}[i]~\mbox{is a top-}n_{adv} ~\mbox{largest value in }\bm{v}\}$
	\end{algorithmic}
    \end{minipage}
    }
\end{algorithm}
\vspace{-10pt}
Once receiving the reviews, the server can estimate the number of poisoned updates $n_{adv}$ and aggregate the rankings. Since some reviewers may overestimate or underestimate $n_{adv}$, we estimate it by the median of $\{n_{adv}^r | r \in R_t\}$, where $n_{adv}^r$ is the estimated number from reviewer $r$.

To identify poisoned updates among all the model updates, we leverage a simple majority vote mechanism illustrated in Algorithm~\ref{alg:majority_vote} to obtain the indices of the potential poisoned updates. 
\ul{An appealing aspect of the majority vote mechanism is its tolerance for the existence of malicious reviewers who may upload wrong reviews to the server.} In practice, we find that, as long as the number of malicious reviewers is less than the half of the total number of reviewers, FedReview can successfully identify and reject all the poisoned updates in most cases. We note that, if the proportion of the malicious clients is $20\%$, and the number of selected clients in each round is larger than or equal to $10$, then the probability that the number of benign reviewers is larger than or equal to the number of malicious reviewers is over $99\%$.

Similar to FLDetector, FedReview may not have a strict bound w.r.t. time $t$ on the $\ell_2$ norm between the global weights at $t$ and the optimal weights. However, if we approximate FedReview by a biased client selection mechanism guided by reviewers, we could have an ergodic convergence rate for FedReview (similar to Fedbuff \cite{nguyen2022federated}).
We assume the probability that the $k$-th client is selected is $p_k$, and for a benign client $b$ and a malicious client $m$, we have $p_b \gg p_m$ due to our ``review and dispose'' mechanism. With the assumptions in \cite{nguyen2022federated}, we obtain the following ergodic convergence rate for FedReview:
\vspace{-5pt}
\begin{equation}\label{eq:convergence}
    \scalebox{0.8}{$
        \displaystyle
        \frac{1}{T}\sum_{t=0}^{T-1}\|\nabla\ell(\thetab^{t})\|^2  
  \leq \frac{2}{T}[\ell(\thetab^{0}) - \ell(\thetab^{*})] +  E^2_l(2 \tilde\sigma^2 + 2\eps G^2 + LG^2) \frac{1}{T}\sum_{t=0}^{T-1} (\eta_t)^2.
    $}
\end{equation}
%\vspace{-8pt}
where $G$ is the upper bound on the $\ell_2$ norm of the gradients, and $\tilde\sigma^2$ refers to the variance of the stochastic gradients in \cite{li2019convergence, nguyen2022federated}, and $\Pr[\Delta_{tm} \neq 0] \le \eps$ for a small $\eps$.

We note that, as long as $\eta_t \sim O(1/t^c)$, where $c>0$, the limit of the right hand of (\ref{eq:convergence}) is 0 when $T\rightarrow 0$. 
In practice, we find that setting a small learning rate allows FedReview to achieve a well-performing model.

\vspace{-13pt}
\section{Experiments}\label{sec:experiments}
\vspace{-6pt}
\subsection{Experimental Setup}
\vspace{-5pt}

\paragraph{Datasets} We follow~\cite{shejwalkarmanipulating} to use Purchase-100, EMNIST, FEMNIST, and CIFAR-10 for evaluation. For Purchase-100, we randomly select 50000 samples for training and 10000 samples for testing. We randomly divide the training samples into 100 training datasets and allocate them to 100 clients. We measure the global model accuracy on all the testing samples. For EMNIST, the total number of training samples is 112800, which is allocated to 100 clients.

\vspace{-8pt}
\paragraph{Networks} On Purchase-100, we follow~\cite{shejwalkarmanipulating} to employ a multi-layer perceptron network with size $[1024, 1024, 100]$, and the activation function is Tanh function. On EMNIST and FEMNIST, we employ LeNet \cite{lecun1998gradient}, which has three convolutional layers. On CIFAR-10, we use ResNet-18~\cite{He_2016_CVPR}.

\vspace{-8pt}
\paragraph{Federated Learning} We set the number of clients as $100$ for the experiments on Purchase-100, EMNIST, and CIFAR-10. For FEMNIST, the default number of clients is $3597$.  We randomly select $10$ clients in each training round for the experiments on Purchase-100, EMNIST, and CIFAR-10. For FEMNIST, we randomly select $30$ clients in each round. We employ an SGD optimizer and set the batch size as 32 for local training. For Purchase-100, EMNIST, and FEMNIST, we set the learning rate as $0.01$ and the momentum as $0.9$. For CIFAR-10, we set the learning rate as $0.01$. In each round, the selected clients train local models for $5$ epochs and then upload the models to the server. The total number of training rounds is set to $100$.

\vspace{-10pt}
\paragraph{Attack Settings} By default, we follow \cite{shejwalkarmanipulating, fang2020local} to set the number of malicious clients as $20\%$ of the total number of the clients. For the min-max and min-sum attacks, we follow \cite{shejwalkarmanipulating} to set $\gamma_{init}$ as $50$ and $\tau$ as $10^{-5}$. For the adaptive attack, we increase $\tau$ to accelerate convergence, otherwise, the adaptive attack will be very slow since it needs multiple surrogate reviewers to evaluate the malicious update in each iteration. \ul{For the experiments of FedReview, if a selected reviewer is a malicious client, the reviewer will upload a wrong ranking, where the ranks of the malicious updates are lower than the ranks of benign updates.}

\vspace{-10pt}
\paragraph{Defense Settings} We set the number of reviewers as the number of selected clients for local training. We set $k$ in Algorithm~\ref{alg:adv_estimation} as $1$. We additionally find that the performance of FedReview is not sensitive to the change $k$ when the proportion of malicious clients is $20\%$.
% If we increase $k$ to $2$, then Algorithm~\ref{alg:adv_estimation} will underestimates the number of poisoned updates in some cases, especially when the standard deviation of the loss is small. 
For the robust aggregation methods, we follow the default settings in the previous works \cite{fang2020local, shejwalkarmanipulating}.

\vspace{-10pt}
\subsection{Main Results}
\vspace{-6pt}
\begin{table}
\vspace{-15pt}
	\begin{center}
    \caption{Compare the performance of different defenses (without access to data) under the scaling model poisoning attack.} \label{tab:defense}
    \vspace{-5pt}
		\scalebox{0.77}{
			\begin{tabular}{ccccccc}
				\hline
				\multirow{2}{*}{Dataset} & \multicolumn{3}{c}{Purchase-100} & EMNIST & CIFAR-10 & FEMNIST \\
				& $\lambda=5.0$ & $\lambda=4.0$ & $\lambda=3.0$ & $\lambda=5.0$ & $\lambda=5.0$ &  $\lambda=5.0$ \\
				\hline
				FedAvg (No Defense) & $0.35\%$ & $18.44\%$ & $62.33\%$ & $2.13\%$  & $10.00\%$ & $3.41\%$ \\
				M-Krum & $14.26\%$ & $27.51\%$ & $26.27\%$ & $2.20\%$ & $10.00\%$ & $38.75\%$ \\
				Median & $37.59\%$ & $42.16\%$ & $48.60\%$ &$30.20\%$ & $10.08\%$ & $49.08\%$ \\
				Trimmed-Mean & $52.14\%$ & $57.75\%$ & $63.18\%$ & $48.34\%$ &  $11.29\%$ & $23.16\%$ \\
				ARFED & $49.96\%$ & $62.89\%$ & $69.34\%$ & $62.78\%$ &  $66.19\%$ & $51.43\%$ \\
				\rowcolor{Gray} FedReview & $82.98\%$ &  $82.33\%$ & $83.95\%$ & $78.70\%$ & $84.56\%$ & $56.54\%$ \\
				\hline
		\end{tabular}}
	\end{center}
\end{table}
\vspace{-20pt}

\vspace{-10pt}
\begin{wrapfigure}{R}{0.43\textwidth}
	\centering
    \vspace{-0.3cm}
    \includegraphics[width=\linewidth]{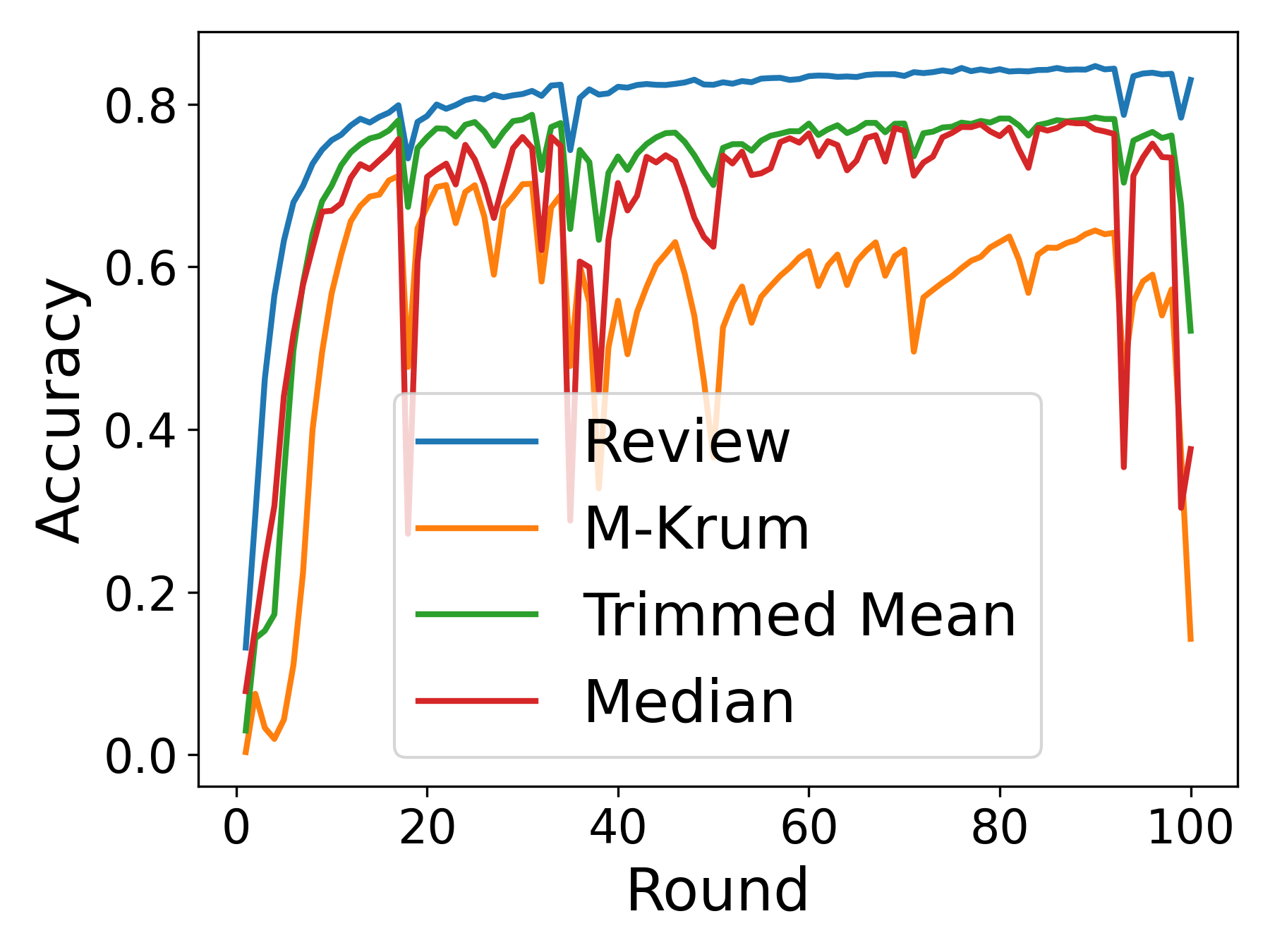}
    \vspace{-1cm}
    \caption{Compare our review mechanism and robust aggregation methods under the scaling model poisoning attack with $\lambda=5.0$.}
    \label{fig:purchase_defenses}
    \vspace{-0.5cm}
\end{wrapfigure}
We compare the performance of different methods against the scaling model poisoning attack in Table~\ref{tab:defense}, where ARFED \cite{isik2023arfed} is a recent effective defense against model poisoning.
As shown in Table~\ref{tab:defense}, our FedReview achieves the best model accuracy among all the methods. When $\lambda$ is small ({\em e.g.,} $\lambda=3.0$), the negative effects of the scaled poisoned updates are mild. Thus, FedAvg can achieve over $60\%$ model accuracy on Purchase-100 against the model poisoning attack. In this case, the robust aggregation methods still eliminate some elements in the benign updates, leading to performance degradation. Therefore, FedAvg can achieve better performance than those robust aggregation methods when $\lambda$ is small.

%\vspace{-2pt}
When we increase $\lambda$ to $5$, the accuracy of the model trained by FedAvg without defense is similar to the accuracy of random guessing, which means the FedAvg completely loses its utility under the scaling attack. But FedReview exhibits strong resistance against the attack (Fig.~\ref{fig:purchase_defenses}). Compared to FedAvg in a benign environment (Table~\ref{tab:no_attack}), the model accuracy achieved by FedReview under the scaling attack only drops by $2\%\sim 6\%$ on all datasets. Since FEMNIST and Purchase-100 has more than 50 classes, this accuracy drop is acceptable.

\begin{wraptable}{L}{0.48\textwidth}
\vspace{-25pt} 
    \centering
    \caption{Compare different methods \textbf{\underline{under no attack}}.} 
    \label{tab:no_attack}
    \resizebox{\linewidth}{!}{
        \begin{tabular}{ccccc}
            \hline
            Dataset & Purchase-100 & EMNIST & CIFAR-10 & FEMNIST \\
            \hline
            FedAvg & $85.45\%$ & $81.06\%$  & $87.76\%$ & $62.46\%$ \\
            M-Krum & $81.67\%$ & $79.46\%$ & $85.93\%$ & $54.98\%$ \\
            Median & $85.29\%$ & $78.28\%$  & $85.60\%$ & $62.33\%$\\
            Trimmed-Mean & $84.72\%$ & $79.37\%$ &  $87.72\%$ & $61.17\%$ \\
            \rowcolor{Gray}FedReview & $85.70\%$ & $80.45\%$ & $87.80\%$ & $62.56\%$ \\
            \hline
        \end{tabular}
    }
    \vspace{-20pt} 
\end{wraptable}
% 紧接着这行直接写你的正文，千万不要留空行！
We also compare FedReview and other baselines in a benign environment (no adversary) and report the results in Table~\ref{tab:no_attack}. Surprisingly, FedReview can outperform FedAvg in a benign environment in some cases. This is because FedReview can identify and drop low-quality benign updates to improve global accuracy. Compared with FedReview, the robust aggregation methods will cause more performance degradation in a benign environment.
% \textbf{\ul{We further evaluate an adaptive attack that assumes the adversary knows the details of FedReview, and analyze the performance of  FedReview under non-i.i.d. data distributions}}. 
We further evaluate an adaptive attack that assumes the adversary knows the details of FedReview and analyze FedReview under non-i.i.d. data distributions. FedReview substantially limits the adaptive attack's negative effects, while the non-i.i.d. variant maintains robustness, with only a 2\%--4\% accuracy drop from benign FedAvg in most settings.
Additionally, we conduct experiments that allow $10\%$ clients to drop out from one round and rejoin the next round ({\em i.e.,} the clients for uploading or reviewing updates are selected from the remaining $90\%$ clients). We find that this ``drop out and rejoin'' barely impacts accuracy because FedReview requires no historic knowledge.

\vspace{-12pt}
\section{Conclusion}\label{sec:conclusion}
\vspace{-8pt}
In this paper,  we propose FedReview to review and dispose potential poisoned updates without access to validation datasets or historic knowledge. FedReview randomly selects a subset of clients as reviewers to compute the loss of the model updates on their training datasets. Based on the loss of the updates, the reviewers estimate the number of poisoned updates by the number of large loss outliers and rank the model updates. A majority vote mechanism is applied to aggregate rankings to reduce the effects of wrong rankings from malicious clients. Extensive evaluations demonstrate that our defense nearly eliminate the negative effects caused by poisoned updates.

\vspace{-8pt}

\bibliographystyle{splncs04}
\bibliography{mybibliography}

\end{document}